\documentclass[11pt,a4paper]{article}
\usepackage[hyperref]{emnlp2018}
\usepackage{times}
\usepackage{latexsym}

\usepackage{csquotes}

\usepackage{graphicx}  
\usepackage{algorithm}
\usepackage[noend]{algpseudocode}
\usepackage{amsmath}
\usepackage{amsfonts}
\usepackage{amssymb}

\usepackage{url}

\aclfinalcopy 

\title{Event Detection with Neural Networks: A Rigorous Empirical Evaluation}

\author{J. Walker Orr \footnotemark[1]\\
George Fox University\\
414 N Meridian St.\\
Newberg, OR 97132\\
{\tt jorr@georgefox.edu}
\\\And
Prasad Tadepalli {\normalfont and} Xiaoli Fern\\
Oregon State University\\
Corvallis OR 97331\\
{\tt tadepall@eecs.oregonstate.edu,} \\
{\tt xfern@eecs.oregonstate.edu}
}

\date{}

\begin{document}
\maketitle
\begin{abstract}
  Detecting events and classifying them into predefined types is an important step in knowledge extraction from natural language texts.
   While the 
  neural network models have generally led the state-of-the-art, the differences in  performance between different architectures have not been rigorously studied. In this paper we present a novel GRU-based model that combines syntactic information along with temporal structure through an attention mechanism. We show that it is competitive with other neural network architectures  through empirical evaluations under different random initializations and training-validation-test splits of ACE2005 dataset.   
   
\end{abstract}

\section{Introduction}

\footnotetext[1]{Also associated with Oregon State University.}
Events are the lingua franca of news stories and narratives and describe important 
changes of state in the world.
Identifying events and classifying them into different types is 
a challenging aspect 
of understanding text. 
This paper focuses on the task of {\em event detection}, which includes identifying the 
``trigger" words that indicate events and classifying the events into refined 
types. Event detection is the necessary first step in inferring more semantic 
information about the events including extracting the arguments of events 
and recognizing temporal and causal relationships between different events.

Neural network models have been the most successful methods for event detection.
However, most current models ignore the syntactic relationships in the text. One of the main contributions of our work is a new DAG-GRU architecture \citep{chung2014empirical} that 
captures the context and syntactic information through 
a bidirectional reading of the text with dependency parse relationships.
This generalizes the GRU model to operate on a graph by novel use of an attention mechanism.




Following the long history of prior work on event detection, ACE2005 is used for the precise definition of the task and the data for the purposes of evaluation. 
One of the challenges of the task is the size and sparsity of this dataset. It consists of 599 documents, which are  broken into a training, development, testing split of 529, 30, and 40 respectively. This split has become a 
de-facto evaluation standard since \cite{li2013joint}.
Furthermore, the test set is small and consists only of newswire documents, when there are multiple domains within ACE2005.
These two factors lead to a significant difference between the training and testing event type distribution.
Though some work had been done comparing method across domains \cite{nguyen2015event}, variations in the training/test split including all the domains has not been studied. We evaluate 
the sensitivity of model accuracy to changes in training and test set splits 
through a randomized study. 

Given the limited amount of training data in  comparison to other datasets used by neural network models, and the narrow margin between many high performance methods, the effect of the initialization of these methods needs to be considered. 
In this paper, we conduct an empirical study of the sensitivity of the system performance to the model initialization. 

Results show that our DAG-GRU method 
is competitive with other state-of-the-art methods. However, 
the performance of all methods is more sensitive to the random model initialization 
than expected. 
Importantly, the ranking of different methods based on the performance on the standard training-validation-test split is sometimes 
different from the ranking based on the average over multiple splits, suggesting that the community should move 
away from single split evaluations. 

\section{Related Work}

Event detection and extraction are well-studied tasks with a long history of research.

\citet{nguyen2015event} used CNNs to represent windows around candidate triggers.
Each word is represented by a concatenation of its word and entity type embeddings with the distance to candidate trigger.
Global max-pooling summarizes the CNN filter and the result is passed to a linear classifier.

\citet{nguyen2016modeling} followed up with a skip-gram based CNN model
which allows the filter to 
skip non-salient or otherwise unnecessary words in the middle of word sequences.

\citet{feng2016language} 
combined a CNN, similar to 
\cite{nguyen2015event}, with
a bi-directional LSTM \cite{hochreiter1997long}
to create a hybrid network.
The output of both networks was concatenated together and fed to a linear model for final predictions.

\citet{nguyen2016joint} uses a bidirectional gated recurrent units (GRUs) for sentence level encoding, and in conjunction with a memory network, to jointly predict events and their arguments. 

\citet{liu2016leveraging} created a probablistic soft logic model incorporating the semantic 
frames from Framenet \citep{baker1998berkeley} in the form of extra training examples. 
Based on the intuition that entity and argument information is important for event detection, \citet{liu2017exploiting} built an attention model over annotated arguments and the context surrounding event trigger candidates.

\citet{liu2018event} created a cross language attention model for event detection and used it for event detection in both the English and Chinese portions of the ACE2005 data.


Recently, \citet{nguyen2018graph} used graph-CNN (GCCN) 
where the convolutional filters are applied to syntactically dependent words in addition to consecutive words.
The addition of the entity information into the network structure produced the state-of-the-art CNN model.

Another neural network model that includes syntactic dependency relationships is DAG-based LSTM \citep{qian2018}.
It combines the syntactic hidden vectors by weighted average and
adds them through a dependency gate to the output gate of the LSTM model.
To the best of our knowledge, none of the neural models combine syntactic information with attention, which motivates our research. 

\section{DAG GRU Model}
Event detection is often framed as a multi-class classification problem \citep{chen2015event,ghaeini2016event}. 
The task is to predict the event label for each word in the test documents, and 
\textit{NIL} if the word is not an event. 
A sentence is a sequence of words $x_1 \dots x_n$, where each word is represented by a $k$-length vector.
The standard GRU model works as follows: \\
 \\
 \hspace*{0.4in}$r_t = \sigma(W_r x_t +  U_r h_{t -1} + b_r)$ \\
 \hspace*{0.4in}$z_t = \sigma(W_z x_t +  U_z h_{t -1} + b_z)$ \\
 \hspace*{0.4in}$\tilde{h}_t = \text{tanh}(W_h x_t + r_t \odot  U_h h_{t -1} + b_h)$\\
 \hspace*{0.4in}$h_t = (1 - z_t) \odot h_{t-1} + z_t \odot \tilde{h}_t$\\
 
The GRU model produces a hidden vector $h_t$ for each word $x_t$ by combining its representation with the previous hidden vector.
Thus $h_t$ summarizes both the word and its prior context.  

Our proposed DAG-GRU model incorporates syntactic information through dependency parse relationships and is 
similar in spirit to \cite{nguyen2018graph} and \cite{qian2018}. 
However, unlike those methods, DAG-GRU uses attention to combine syntactic and temporal information.
Rather than using an additional gate as in \cite{qian2018}, DAG-GRU creates a single combined representation over previous hidden vectors and then applies the standard GRU model.
Each relationship is represented as an edge, $(t,t^\prime, e)$, between words at index $t$ and $t^\prime$ with an edge type $e$.
The standard GRU edges are included as $(t, t-1, temporal)$. 

\begin{figure*}
\centering
\includegraphics[scale=.75]{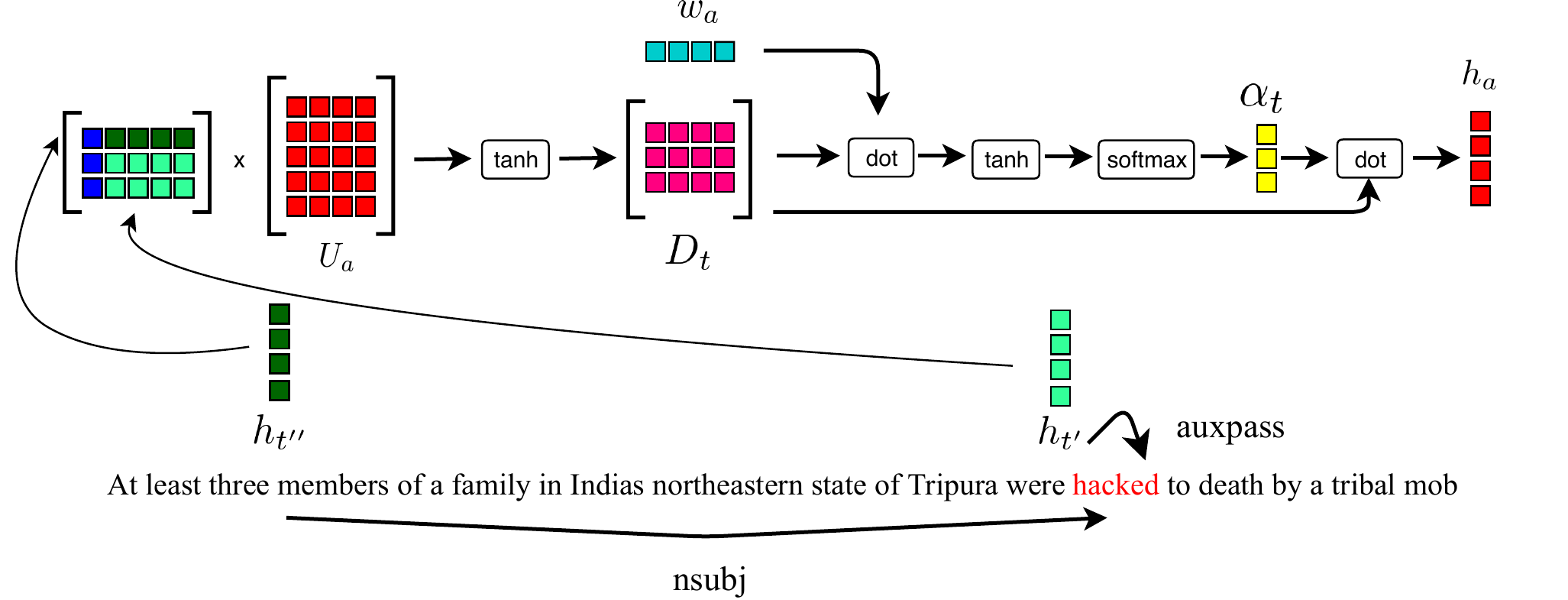}
\caption{The hidden state of ``hacked'' is a combination of previous output vectors. In this case, three vectors are aggregated with DAG-GRU's attention model. $h_{t^{\prime\prime}}$, is included in the input for the attention model since it is accessible through the ``subj'' dependency edge. $h_{t^\prime}$ is included twice because it is connected through a narrative edge and a dependency edge with type ``auxpass.'' The input matrix is non-linearly transformed by $U_a$ and $\text{tanh}$. Next, $w_a$ determines the importance of each vector in $D_t$. Finally, the attention $a_t$ is produced by tanh followed by softmax then applied to $D_t$. The subject ``members'' would be distant under a standard RNN model, however the DAG-GRU model can focus on this important connection via dependency edges and attention.}
\end{figure*}

Each dependency relationship may be between any two words, which could produce a graph with cycles. 
However, back-propagation through time \citep{mozer1995focused} requires a directed acyclic graph (DAG), 
Hence the sentence graph, consisting of temporal and dependency edges $E$, is split into two DAGs:
a ``forward'' DAG $G_f$ that consists of only of edges $(t, t^\prime, e)$ where $t^\prime < t$ and a corresponding ``backward'' DAG $G_b$ where $t^\prime > t$.
The dependency relation between $t$ and $t^\prime$ also includes the parent-child orientation, e.g., $nsubj\text{-}parent$ or $nsubj\text{-}child$ for a $nsubj$ (subject) relation.


An attention mechanism is used to combine the multiple hidden vectors. 
The matrix $D_t$ is formed at each word $x_t$ by collecting and transforming all the previous hidden vectors coming into node $t$, one per each edge type $e$. 
$\alpha$ gives the attention, a distribution weighting importance over the edges.
Finally, the combined hidden vector $h_a$ is created by summing $D_t$ weighted by attention.
\begin{align}
&D_t^T =[ {\text{tanh}}(U_e h_{t^\prime})| (t, t^\prime, e) \in E] \nonumber \\
&\alpha = \text{softmax}(\text{tanh}(D_t w_a)) \nonumber \\
&h_a = D_t^T \alpha \nonumber
\end{align}

However, having a set of parameters $U_e$ for each edge type $e$ is over-specific for small datasets. 
Instead a shared set of parameters $U_a$ is used in conjunction with an edge embedding $v_e$.
\begin{align*}
&D_t^T =[ {\text{tanh}}(U_a h_{t^\prime} \circ v_e)| (t, t^\prime, e) \in E]
\end{align*}

The edge type embedding $v_e$ is concatenated with the hidden vector $h_{t^\prime}$ and then transformed by the shared weights $U_a$.
This limits the number of parameters while flexibly weighting the different edge types. 
The new combined hidden vector $h_a$ is used instead of $h_{t-1}$ in the GRU equations.


The model is run forward and backward with the output concatenated, $h_{c,t} = h_{f,t} \odot h_{b,t}$, for a representation that includes the entire sentence's context and dependency relations.
After applying dropout \citep{srivastava2014dropout} with $0.5$ rate to $h_{c,t}$, a linear model with softmax is used to make predictions for each word at index $t$.

\section{Experiments}

We use the ACE2005 dataset for evaluation. 
Each word in each document is marked with one of the thirty-three event types or \textit{Nil} for non-triggers.
Several high-performance models were reproduced for comparison.
Each is a good faith reproduction of the original with some adjustments to 
level the playing field. 

For word embeddings, Elmo was used to generate a fixed representation for every word in ACE2005 \citep{peters2018deep}.
The three vectors produced per word were concatenated together for a single representation.
We did not use entity type embeddings for any method.
The models were trained to minimize the cross entropy loss with Adam \citep{kingma2014adam} with L2 regularization set at $0.0001$.
The learning rate was halved every five epochs starting from $0.0005$ for a maximum of $30$ epochs or until convergence as determined by F1 score on the development set.

The same training method and word embeddings were used across all the methods.
Based on preliminary experiments, these settings resulted in better performance than those originally specified.
However, notably both GRU \citep{nguyen2016joint} and DAG-LSTM \citep{qian2018} were not used as joint models.
Further, the GRU implementation did not use a memory network, instead we used the final vectors from the forward and backward pass concatenated to each timestep's output for additional context.
For CNNs \citep{nguyen2015event} the number of filters was reduced to 50 per filter size.
The CNN+LSTM \citep{feng2016language} model had no other modifications.
DAG-GRU used a hidden layer size of $128$.
Variant A of the DAG-GRU model utilized the attention mechanism, while variant B used averaging, that is:
\begin{align*}
D_t = \frac{1}{|E(t)|} \underset{(t^\prime, e) \in E(t)}{\sum} \text{tanh}(U_a h_{t^\prime}\circ v_e)
\end{align*}

\subsection{Effects of Random Initialization}

\begin{figure}
\includegraphics[scale=.35]{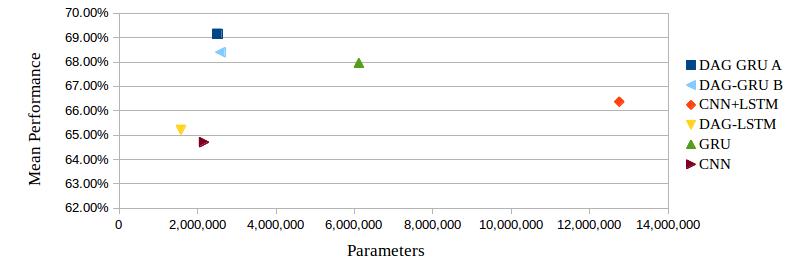}
\caption{A comparison of mean performance versus number of parameters.}
\end{figure}

Given that ACE2005 is small as far as neural network models are concerned, 
the effect of the random initialization of these models needs to be studied.
Although some methods include tests of significance, the type of statistical test is often not reported.
Simple statistical significance tests, such as the t-test, are not compatible with a single F1 score, instead the average of F1 scores should be tested \citep{f1confidence}.

We reproduced and evaluated five different systems with different initializations to empirically assess the effect of initialization.
The experiments were done on the standard ACE2005 split and the aggregated results over $20$ random seeds were given in Table \ref{seed}.
The random initializations of the models had a significant impact on their performance.
The variation was large enough that the observed range of the F1 scores overlapped across almost all the models.
However the differences in average performances of different methods, except for CNN and DAG-LSTM, were significant at $p < 0.05$ according to the t-test, not controlling for multiple hypotheses.

Both the GRU \citep{nguyen2016joint} and CNN \citep{nguyen2015event} models perform well with their best scores being close to the reported values.
The CNN+LSTM model's results were significantly lower than the published values, though this method has the highest variation.
It is possible that there is some unknown factor such as the preprocessing of the data that significantly impacted the results or that the value is an outlier.
Likewise, the DAG-LSTM model underperformed. However, the published results were based on a joint event and argument extraction model
and 
probably benefited from the additional entity and argument information. 

DAG-GRU A consistently and significantly outperforms the other methods in this comparison.
The best observed F1 score, 71.1\%, for DAG-GRU is close to the 
published state-of-the-art scores of DAG-LSTM and GCNN at 
71.9\% and 71.4\% respectively. 
With additional entity information, GCNN achieves a score of 73.1\%.
Also, the attention mechanism used in DAG-GRU A shows a significant improvement over the averaging method of DAG-GRU B.
This indicates that some syntactic links are more useful than others and that the weighting attention applies is necessary to utilize that syntactic information.

Another source of variation was the distributional differences between the development and testing sets.
Further, the testing set only include newswire articles whereas the training and dev. sets contain informal writing such as web log (WL) documents.
The two sets have different proportions of event types and each model saw at least a 2\% drop in performance between dev. and test on average.
At worst, the DAG-LSTM model's drop was 5.26\%.
This is a problem for model selection, since the dev. score is used to choose the best model, hyperparameters, or random initialization.
The distributional differences mean that methods which outperform others on the dev. set do not necessarily perform as well on the test set.
For example, DAG-GRU A performs worse that DAG-GRU B on the dev. set, however it achieves a higher mean score on the testing set.

One method of model selection over random initializations is to train the model $k$ times and pick the best one based on the dev. score.
Repeating this model selection procedure many times for each model is prohibitively expensive, so the experiment was approximated by bootstrapping the twenty samples per model \citep{efron1992bootstrap}.
For each model, 5 dev. \& test score pairs were sampled with replacement from the twenty available pairs.
The initialization with the best dev. score was selected and the corresponding test score was taken.
This model selection process of picking the best of 5 random samples was repeated 1000 times and the results are shown in Table \ref{bootstrap}.
This process did not substantially increase the average performance
beyond the results in Table~\ref{seed}, although it did reduce the variance, except for the CNN model.
It appears that using the dev. score for model selection is only marginally helpful.

\begin{table}
\tiny
\centering
\setlength\tabcolsep{3pt}
\begin{tabular}{|l | l | l | l | l | l | l |}
\hline
\textbf{Model} & \textbf{Dev Mean} & \textbf{Mean} & \textbf{Min} & \textbf{Max} & \textbf{Std. Dev.} & \textbf{Published}\\
\hline
DAG-GRU A & 70.3\% & 69.2\% $\pm$ 0.42 & 67.8\% & 71.1\% & 0.91\% & -\\
DAG-GRU B & 71.2\% & 68.4\% $\pm$ 0.45 & 67.39\% & 70.53\% & 0.96\% & - \\ 
GRU & 70.3\% & 68.0\% $\pm$ 0.42 & 66.4\% & 69.4\% & 0.86\% & 69.3\%$\dagger$\\
CNN+LSTM & 69.6\% & 66.4\% $\pm$ 0.62 & 63.6\% & 68.1\% & 1.32\% & 73.4\%\\
DAG-LSTM & 70.5\% & 65.2\% $\pm$ 0.44 & 63.0\% & 66.8\% & 0.94\% & 71.9\%$\dagger$\\
CNN & 68.5\% & 64.7\% $\pm$ 0.65 & 61.6\% & 67.2\% & 1.38\% & 67.6\%\\
\hline
\end{tabular}
\caption{The statistics of the 20 random initializations experiment. $\dagger$ denotes results are from a joint event and argument extraction model.}
\label{seed}
\end{table}

\begin{table}
\tiny
\centering
\begin{tabular}{| l | l | l | l | l | l |}
\hline
\textbf{Model} & \textbf{Dev Mean} & \textbf{Mean} & \textbf{Std. Dev.}\\
\hline
DAG-GRU A & 72.0\% & 69.2\% $\pm$ 0.04\% & 0.68\%\\
DAG-GRU B & 72.0\% & 67.9\% $\pm$ 0.04\% & 0.60\%\\
GRU & 71.5\% & 68.4\% $\pm$ 0.05\% & 0.80\%\\
CNN+LSTM & 70.8\% & 66.8\%  $\pm$ 0.07\% & 1.08\%\\
DAG-LSTM & 70.4\% & 65.5\%  $\pm$ 0.02\% & 0.40\%\\
CNN & 69.6\% & 65.4\% $\pm$ 0.09\% & 1.49\%\\
\hline
\end{tabular}
\caption{Bootstrap estimates on 1000 samples for each model after model selection based on dev set scores.}
\label{bootstrap}
\end{table}

\subsection{Randomized Splits}
In order to explore the effect of the training/testing split popularized by \cite{li2013joint}, a randomized cross validation experiment was conducted.
From the set of 599 documents in ACE2005, 10 random splits were created maintaining the same 529, 30, 40 document counts per split, training, development, testing, respectively. 
This method was used to evaluate the effect of the standard split, since it maintains the same data proportions while only varying the split.
The results of the experiment are found in Table \ref{cross}.

\begin{table}
\centering
\setlength\tabcolsep{3pt}
\tiny
\begin{tabular}{| l | l | l | l | l | l |}
\hline
\textbf{Method} & \textbf{Dev Mean} & \textbf{Mean} & \textbf{Min} & \textbf{Max} & \textbf{Std. Dev.}\\
\hline
DAG-GRU A & 71.4\% & 68.4\% $\pm$ 1.85\% & 65.7\% & 74.1\% & 2.59\% \\
DAG-GRU B & 70.9\% & 68.4\% $\pm$ 1.88\% & 64.19\% & 73.59 & 2.63\% \\
DAG-LSTM & 68.9\% & 67.3\% $\pm$ 1.43\% & 63.5\% & 70.7\% & 2.00\% \\
GRU & 69.8\% & 66.6\% $\pm$ 1.86\% & 62.5\% & 71.1\% & 2.60\% \\
CNN+LSTM & 69.8\% & 66.3\% $\pm$ 2.03\% & 60.1\% & 70.3\% & 2.83\% \\
CNN & 68.0\% & 65.4\% $\pm$ 1.59\% & 60.7\% & 69.2\% & 2.22\% \\
\hline
\end{tabular}
\caption{Average results on 10 randomized splits.}
\label{cross}
\end{table}


The effect of the split is substantial.
Almost all models' performance dropped except for DAG-LSTM, however the variance increased across all models.
In the worst case, the standard deviation increased threefold from 0.86\% to 2.60\% for the GRU model.
In fact, the increased variation of the splits means that the confidence intervals for all the models overlap.
This aligns with cross domain analysis, some domains such as WL are known to be much more difficult than the newswire domain which comprises all of the test data under the standard split \citep{nguyen2015event}.
Further, the effect of the difference in splits also negates the benefits of the attention mechanism of DAG-GRU A.
This is likely due to the test partitions' inclusion of WL and other kinds of informal writing.
The syntactic links are much more likely to be noisy for informal writing, reducing the syntactic information's usefulness and reliability.

All these sources of variation are greater than most advances in event detection, so quantifying and reporting this variation is essential when assessing model performance.
Further, understanding this variation is important for reproducibility and is necessary for making any valid claims about a model's relative effectiveness. 

\section{Conclusions}

We introduced and evaluated a 
DAG-GRU model along with four previous models in two different settings, the standard ACE2005 split with multiple random initializations and 
the same dataset with multiple random splits.
These experiments demonstrate that our model, which utilizes syntactic information through an attention mechanism, is competitive with the state-of-the-art.
Further, they show that there are several significant sources of variation which had not been previously studied and quantified. 
Studying and mitigating this variation could be of significant value by itself. 
At a minimum, it suggests that the community should move away from evaluations based on single random initializations and single training-test splits. 

\section*{Acknowledgments}
This work was supported by grants from DARPA XAI (N66001-17-2-4030), DEFT (FA8750-13-2-0033), and NSF (IIS-1619433 and IIS-1406049).

\bibliography{events}
\bibliographystyle{acl_natbib_nourl}

\end{document}